\documentclass[letterpaper]{article} % DO NOT CHANGE THIS
\usepackage{aaai23}  % DO NOT CHANGE THIS
\usepackage{times}  % DO NOT CHANGE THIS
\usepackage{helvet}  % DO NOT CHANGE THIS
\usepackage{courier}  % DO NOT CHANGE THIS
\usepackage[hyphens]{url}  % DO NOT CHANGE THIS
\usepackage{graphicx} % DO NOT CHANGE THIS
\usepackage{natbib}  % DO NOT CHANGE THIS AND DO NOT ADD ANY OPTIONS TO IT
\usepackage{caption} % DO NOT CHANGE THIS AND DO NOT ADD ANY OPTIONS TO IT
\usepackage{algorithm}
\usepackage{algorithmic}
\usepackage{amsmath}
\usepackage{multirow}
\usepackage[dvipsnames]{xcolor}
\usepackage{subcaption}
\usepackage{graphicx}

\usepackage{newfloat}
\usepackage{listings}
\DeclareCaptionStyle{ruled}{labelfont=normalfont,labelsep=colon,strut=off} % DO NOT CHANGE THIS
\floatstyle{ruled}
\newfloat{listing}{tb}{lst}{}
\floatname{listing}{Listing}
\title{Astromorphic Self-Repair of Neuromorphic Hardware Systems}
\author{
Zhuangyu Han,
A N M Nafiul Islam,
Abhronil Sengupta
}
\affiliations{
School of Electrical Engineering and Computer Science\\
Penn State University\\
University Park, PA 16802 \\
zfh5141@psu.edu, nafiul@psu.edu, sengupta@psu.edu
}

\begin{document}

\maketitle

\begin{abstract}
While neuromorphic computing architectures based on Spiking Neural Networks (SNNs) are increasingly gaining interest as a pathway toward bio-plausible machine learning, attention is still focused on computational units like the neuron and synapse. Shifting from this neuro-synaptic perspective, this paper attempts to explore the self-repair role of glial cells, in particular, astrocytes. The work investigates stronger correlations with astrocyte computational neuroscience models to develop macro-models with a higher degree of bio-fidelity that accurately captures the dynamic behavior of the self-repair process. Hardware-software co-design analysis reveals that bio-morphic astrocytic regulation has the potential to self-repair hardware realistic faults in neuromorphic hardware systems with significantly better accuracy and repair convergence for unsupervised learning tasks on the MNIST and F-MNIST datasets. Our implementation source code and trained models are available at 
\url{https://github.com/NeuroCompLab-psu/Astromorphic_Self_Repair}.
\end{abstract}

\section{Introduction}
As a pathway to enable next-generation neuromorphic intelligence, Spiking Neural Networks (SNNs) are emerging as a disruptive computing paradigm where the information transmitted and processed in the artificial neurons of the computing system is embodied in a temporal series of binary spikes which mimic the action potentials propagating between neurons in the mammalian brain. SNNs have demonstrated significant advantages in comparison to traditional non-spiking computational architectures in several aspects, such as low latency and low energy consumption \cite{davies_loihi_2018, merolla_million_2014, sengupta_encoding_2017, diehl_unsupervised_2015, lu_exploring_2020, neftci_surrogate_2019}. 

Recent research in neuromorphic computing has started focusing on other cellular components in the brain that might contribute to cognition in addition to neuronal spiking behavior and synaptic plasticity. Specifically, this work explores the contribution of one such component - glial cells, in particular, the astrocytes \cite{oberheim_astrocytic_2006}. It has been observed that neurotransmitters released from a neuron can trigger diffusing $\mathrm{Ca}^{2+}$ waves inside the cytoplasm of neighboring astrocytes, which later regulates the strength of the synapses surrounded by these astrocytes \cite{cornell-bell_glutamate_1990, perea_astrocytes_2007}. Previous works have considered the regulation mechanism of the astrocytes as a self-repair capability of artificial spiking neural networks \cite{wade_self-repair_2012, liu2017spanner,liu2018exploring,rastogi_self-repair_2021} with synaptic faults. This is especially timely given the significant advancements in Artificial Intelligence (AI) memristive hardware over the past few years enabled by devices where the core physics serve as a natural hardware substrate for designing compact, energy-efficient neuromorphic hardware accelerators \cite{sung2018perspective,park2022complex}. However, such devices often suffer from non-idealities and faults \cite{liu_online_2014, radetzki_methods_2013} and therefore mitigation of such issues is critical to ensure minimal accuracy degradation in AI hardware platforms. While astrocytic regulation of synaptic transmission probability offers motivation toward autonomous self-repair of faulty neuromorphic hardware, current work significantly falls short of this goal due to a lack of hardware insights. The neuroscience inspired algorithm frameworks are often loosely bio-inspired and depend on \textit{global network level parameters} like synaptic weight percentiles \cite{rastogi_self-repair_2021} which are not practical to calculate in real-time under energy and resource constraints in edge AI systems.

This work forges stronger connections with neuroscience computational models of astrocytes to develop self-repair algorithmic frameworks driven by a software-hardware co-design perspective. The key distinguishing factors of our work against prior proposals are:

\textbf{(i) Neuroscience Inspired Self-Repair Learning Algorithm Formulation:} Based on the bio-inspiration premise that self-repair learning algorithm formulations need to have stronger neuroscience correlation in order to be hardware efficient and compatible, we design a self-repair rule by theoretically analyzing the dynamic temporal behavior of the self-repair process enabled by astrocytes. We develop a macro-model of the self-repair process by benchmarking it to astrocyte computational neuroscience models. As shown in the paper, the bio-inspiration route enables us to construct self-repair-based synaptic learning rules that capture the dynamic temporal repair process while simultaneously depending on \textit{local network level parameters} that make it hardware compatible. Performance evaluation of the proposed learning rule is demonstrated for an unsupervised learning framework on the MNIST and F-MNIST datasets.

\textbf{(ii) Algorithm-Hardware Co-Design:} Going beyond software simulated faults where synapse transmission probabilities (PR) are assumed to be zero, we will analyze the impact of non-idealities occurring in the neuromorphic hardware, which are significantly more complex than stuck-at-zero faults. In particular, we focus on synaptic weight drift effects observed in memristive AI hardware \cite{ambrogio_reducing_2019}, in addition to stuck-at faults. This can therefore lead to the development of highly adaptive neuromorphic systems that can self-repair faults (device level non-idealities) in an autonomous fashion.

\section{Related Work and Main Contributions}
Existing research on AI hardware design with device level faults ranges from fault masking and redundancy repair schemes \cite{chen2012cost} to retraining \cite{xia2017fault}. Both approaches require complex fault detection schemes. In contrast, our repair scheme does not require any fault localization scheme. Further, our self-repair algorithm is entirely local, thereby making it hardware-compatible. Prior retraining based approaches depend on global learning rules. Implementation of local learning rules is an active research area \cite{zenke2021brain}.

In the domain of astromorphic unsupervised self-repair, while prior work like \cite{rastogi_self-repair_2021,liu2017spanner,liu2018exploring} discusses astrocyte computational modelling, they only draw qualitative observations from the model. For instance, the learning algorithm developed in \cite{rastogi_self-repair_2021} is primarily based on the qualitative observation that after the introduction of the faults, the synaptic transmission probability (PR) of the synapses with the higher initial PR value is enhanced greatly compared to the one with the lower initial PR.
On the other hand, our work develops the self-repair algorithm based on quantitative macro-modelling of the astrocyte computational model. This enables the critical modelling of the temporal dynamics of the self-repair process which results in faster repair. 

Another significant contribution of the work is hardware-software co-design where we evaluate the performance of our developed algorithm in presence of hardware non-idealities like conductance drift which is much more complex than prior simulated stuck-at-faults.

Finally, while there have been prior efforts at mimicking astrocyte functionality in hardware \cite{rahiminejad2021neuromorphic}, they primarily focus on implementing the detailed computational model in hardware without considering aspects of astrocyte functionality relevant to self-repair of neuromorphic hardware.

\section{Methods}
\subsection{Astrocyte Mediated Synaptic Dynamics}
Astrocytes modulate the transmission characteristics, fundamentally the spike transmission probability, of the synapses they ensheathe \cite{wade_self-repair_2012}. Endocannabinoids-mediated synaptic potentiation (e-SP) (Navarrete and Araque, 2010) and Depolarization-induced suppression of excitation (DSE) \cite{e_alger_retrograde_2002} are two dynamic modulating factors that influence the PR evolution of synapses. Through a signal pathway inside the astrocyte cytoplasm, e-SP increases the PR after each post-synaptic neuron firing event. In contrast, DSE provides direct negative feedback to the firing rate of post-synaptic neurons. Prior work on computational models of astrocytes has outlined the quantitative change of 2-arachidonoyl glycerol (2-AG) as the first process initiating PR modulation \cite{wade_self-repair_2012}. The post-synaptic neuron releases 2-AG every time it fires, and the 2-AG decays exponentially when there is no release from the dendrite of the post-synaptic neuron. The dynamics of 2-AG can be described as:
\begin{equation} \label{eq1}
\frac{d(\mathrm{AG})}{dt} = \frac{-\mathrm{AG}}{\tau_{\mathrm{AG}}} + r_{\mathrm{AG}}\delta(t-t_{sp})
\end{equation}
where, AG is the quantity of 2-AG released by the post-synaptic neuron, $\tau_{\mathrm{AG}}$ is the decay time constant of 2-AG, $r_{\mathrm{AG}}$ is the production rate of 2-AG and $t_{sp}$ is the time when the post-synaptic neuron spikes. Note that 2-AG is a neuron-specific variable, which implies that the spikes of all synapses of a neuron will contribute to the evolution of 2-AG of this neuron.

The binding process of 2-AG to CB1R receptors on the pre-synaptic neuron triggers DSE. The relationship between 2-AG and DSE can be described by a simple linear equation:
\begin{equation} \label{eq2}
\mathrm{DSE} = -\mathrm{AG} \times K_{\mathrm{AG}}
\end{equation}
where, AG is the quantity of 2-AG released by the post-synaptic neuron and $K_{\mathrm{AG}}$ is a linear scaling factor. DSE is also a neuron-specific variable.

Concurrently, 2-AG also binds to the CB1Rs on the plasma membrane of the astrocyte, which subsequently triggers the generation of $\mathrm{IP}_{3}$ inside the cytoplasm of astrocytes. $\mathrm{IP}_{3}$ later binds on the $\mathrm{IP}_{3}$ receptor in the Endoplasmic Reticulum (ER), which releases $\mathrm{Ca}^{2+}$ into the cytoplasm. The $\mathrm{Ca}^{2+}$ dynamics, described by the Li-Rinzel model \cite{li_equations_1994}, involves three $\mathrm{Ca}^{2+}$ currents which are $J_{\mathrm{chan}}$ (the current through $\mathrm{Ca}^{2+}$ channel controlled by mutual gating of $\mathrm{Ca}^{2+}$ and $\mathrm{IP}_{3}$), $J_{\mathrm{leak}}$ (the leakage current from ER into the cytoplasm) and $J_{\mathrm{pump}}$ (the pumping current where the $\mathrm{Ca}^{2+}$ is absorbed into ER through Sacro-Endoplasmic-Reticulum $\mathrm{Ca}^{2+}$-ATPase (SERCA) pumps). Interested readers are referred to Refs. Wade et al. (2012) and De Pittà et al. (2009) for details on the computational model. The intracellular $\mathrm{Ca}^{2+}$ dynamics can be described as:
\begin{equation} \label{eq3}
\frac{d\mathrm{Ca}^{2+}}{dt} = J_{\mathrm{chan}} + J_{\mathrm{leak}} - J_{\mathrm{pump}} 
\end{equation}
The exocytosis of glutamate is regulated by $\mathrm{Ca}^{2+}$, which can be described as:
\begin{equation} \label{eq4}
\frac{d(\mathrm{Glu})}{dt} = \frac{-\mathrm{Glu}}{\tau_{\mathrm{Glu}}} + r_{\mathrm{Glu}}\delta(t-t_{\mathrm{Ca}^{2+}})
\end{equation}
where, Glu is the total quantity of glutamate released by the astrocyte, $\tau_{\mathrm{Glu}}$ is the glutamate decay time constant, $r_{\mathrm{Glu}}$ is the fixed amount of glutamate release each time the $\mathrm{Ca}^{2+}$ concentration crosses the threshold and $t_{\mathrm{Ca}^{2+}}$ is the time instant when the $\mathrm{Ca}^{2+}$ concentration crosses the threshold.

Glutamate binding is the final stage of potentiation where the synapse has its PR increased when the group I metabotropic Glutamate Receptors (mGluRs) on the membrane of the pre-synaptic neuron receives the glutamate, whose dynamics is described by:
\begin{equation} \label{eq5}
\tau_{\mathrm{eSP}}\frac{d(\mathrm{eSP})}{dt} = -\mathrm{eSP} + m_{\mathrm{eSP}} \mathrm{Glu}(t)
\end{equation}
where, $\mathrm{eSP}$ is the intensity of e-SP, $\tau_{\mathrm{eSP}}$ is the decay time constant of eSP, $m_{\mathrm{eSP}}$ is a scaling factor and $\mathrm{Glu}(t)$ is the released glutamate quantity at time $t$. Please note here that eSP is an astrocyte-specific variable which influences all the synapses of all the neurons ensheathed by the astrocyte.

Potentiation provides positive feedback to synaptic firing, while DSE depresses synaptic firing. The PR dynamics as a function of time can be described as a combined effect of DSE and eSP as follows:
\begin{equation} \label{eq6}
\mathrm{PR}(t) = \mathrm{PR}(0) + \mathrm{PR}(0) \times (\frac{\mathrm{DSE}(t) + \mathrm{eSP}(t)}{100})
\end{equation}
where, $\mathrm{PR}(0)$ is the PR of the synapse before any firing.

As described by the dynamic astrocyte model, DSE is established by the direct binding of 2-AG to the pre-synaptic neuron, which produces a restricted effect on a single neuron. However, the e-SP feedback through the cytoplasm of the astrocyte is globally effective for all synapses ensheathed by the corresponding astrocyte. During the self-repair process enabled by the astrocyte, when the PR of a synapse drops due to faults, the magnitude of DSE of that specific neuron (considering that DSE is always negative) will immediately decline caused by the decreasing 2-AG. However, the global e-SP maintains its intensity due to the continuous firing of the neighboring neurons. Consequently, the PR of healthy synapses for the affected neuron with faults will increase based on Equation \ref{eq6} and recover the spiking activity of the neuron to the initial baseline value. The next section discusses the formulation of a benchmarked astrocyte macro-model that captures the dynamical repair process for hardware realistic faults based on this computational neuroscience model. 

\subsection{Macro-model Formulation for Hardware Realistic Synaptic Fault Repair}

\subsubsection{Neuron-Astrocyte Structure Configuration}
\begin{figure}[h]
\centering
\includegraphics[width=7.8cm]{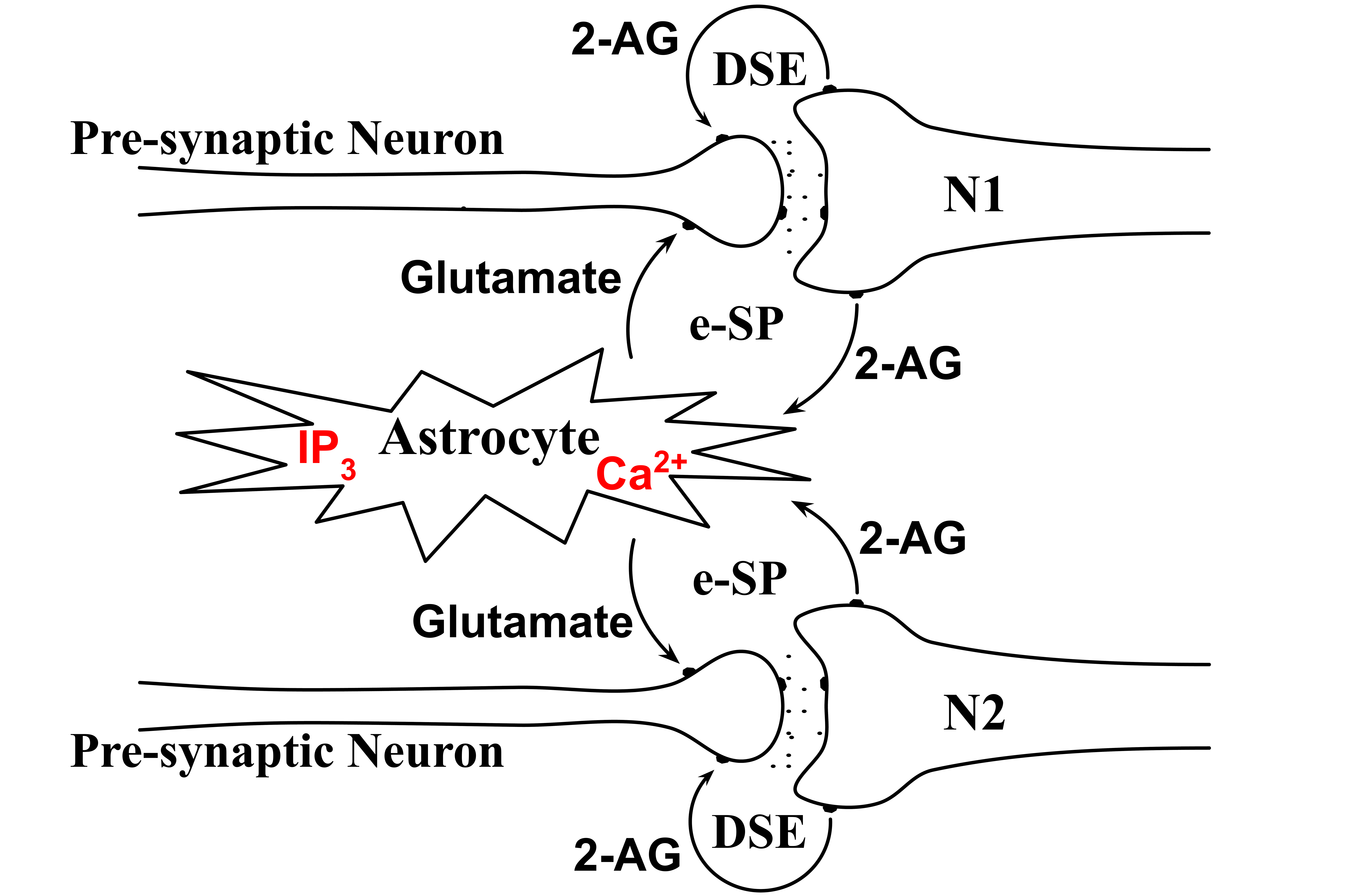}
\caption{Two neurons interacting with a common astrocyte.}
\label{astro-neuron-struct}
\end{figure}
The two-neuron-one-astrocyte computational simulation model \cite{wade_self-repair_2012}, shown in Figure \ref{astro-neuron-struct}, is adopted in this paper. In such a model, there are two neurons, N1 and N2, with identical biological characteristics and one astrocyte functioning as the e-SP pathway for both neurons. The neuron firing model is discussed in detail in the next subsection. The two neurons have the same number of synapses and each synapse receives a Poisson spike train with a globally constant rate parameter, i.e. 10 spikes per second. Please note that Figure \ref{astro-neuron-struct} shows only one dendrite for each neuron, for simplicity, which implies that the DSE shown in the figure is not restricted to the single synapse, instead it is effective neuron-wide. The initial PR values of each synapse are initialized randomly and are sampled from a uniform distribution whose mean is pre-defined. 

The signalling pathway of the astrocyte is computationally realized by a group of global variables, such as the quantity of  $\mathrm{IP}_3$, $\mathrm{Ca}^{2+}$ and glutamate. The interaction between the neurons and the astrocyte is represented by the change of variable 2-AG and glutamate quantity.

The total simulation duration is 400 s and a resolution of 1 ms is used for modelling the temporal dynamics. The Poisson spike trains are continuously received by all the synapses for the entire duration of the simulation. A specific amount of current (6650 pA in this paper) is injected into the neuron if a spike is successfully transmitted through the synapse.

\subsubsection{Neuron Firing Model}
The biological attributes of the neurons are critical to the dynamics of the neuron-astrocyte structure. For the computational simulations performed in this paper, the Leaky-Integrate-Fire model is used for modeling the neuron dynamics. The evolution of neuron membrane potential follows the equation below:
\begin{equation} \label{eq7}
\tau_{v} \frac{d v(t)}{d t} = -(v(t) - v_{\mathrm{res}}) + I(t)
\end{equation}
where, $v$ is the membrane potential, $\tau_{v}$ is the decay time constant of the membrane potential, $v_{\mathrm{res}}$ is the resting state potential and $I(t)$ is the injected current which is the total current the neuron receives from all the synapses. When the membrane potential reaches a threshold potential, $v_{\mathrm{th}}$, the neuron fires once and immediately the membrane potential is set to a pre-determined value, $v_{\mathrm{reset}}$. No refractory period, $\delta_{\mathrm{ref}}$, is considered in this computational model.

\subsubsection{Memristive Hardware Modelling and Fault Simulation}
\begin{figure}[t]
\centering
\includegraphics[width=7.8cm]{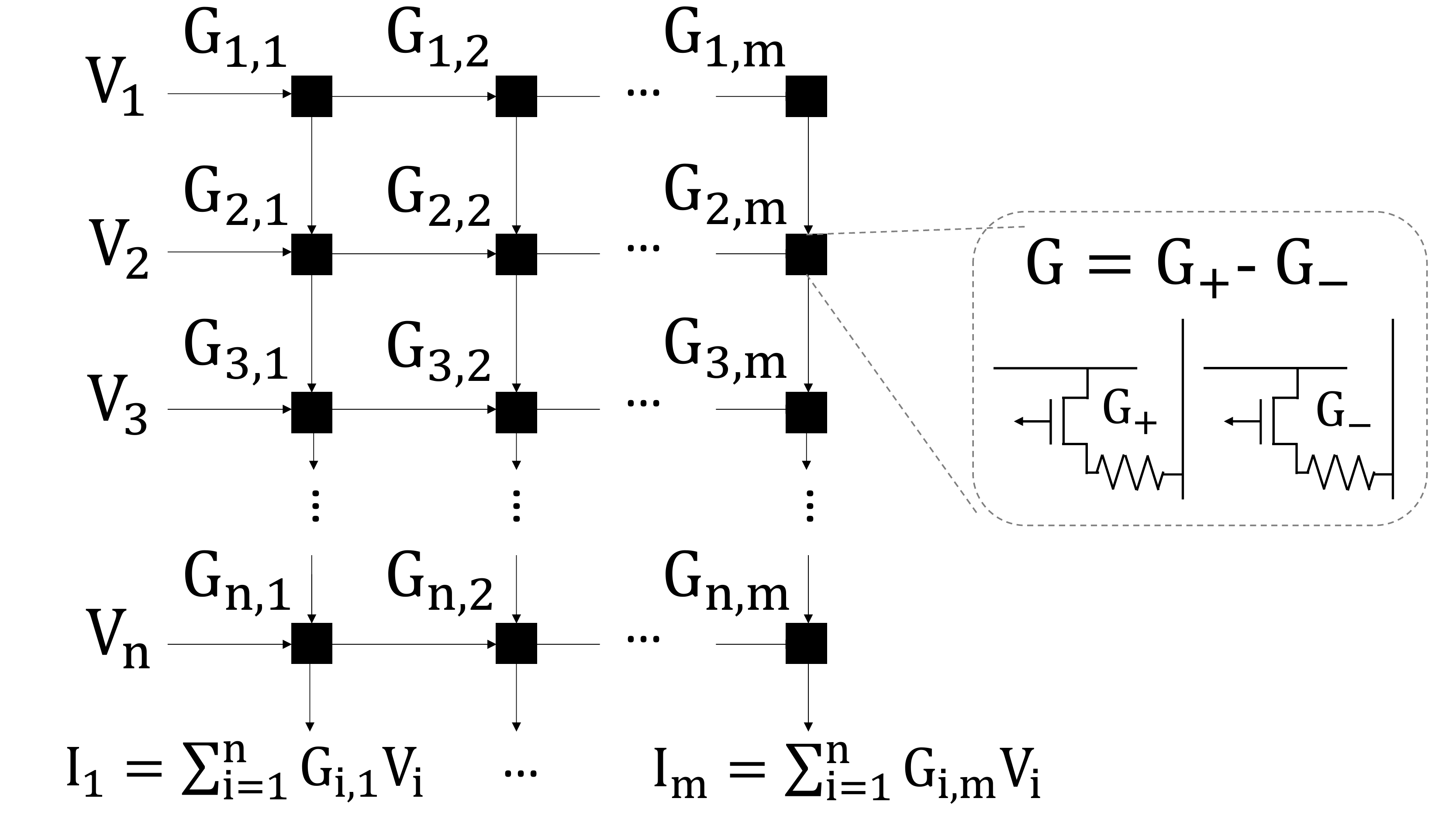}
\caption{Schematic of the network mapped to a memristive crossbar array. Each weight is encoded by the conductance of a pair of PCM devices. Additional peripheral circuitry is omitted.}
\label{crossbar}
\end{figure}
At the hardware level, the PR values (equivalently synaptic weights for artificial spiking neural networks used for machine intelligence) of the network are mapped to the conductance states of the memristive devices. In this paper, we specifically consider device characteristics corresponding to a representative memristive technology - Phase Change Memory (PCM) \cite{fong2017phase}. Organized in a crossbar fashion, as shown in Figure \ref{crossbar}, these arrays of devices can implement the dot product operation. The input voltage $\mathrm{V_i}$ coming from the pre-synaptic neurons to each row of synapses are modulated by the conductance of the synaptic devices and are summed along the columns by Kirchhoff’s current law. Thus, the post-synaptic current fed to the post-synaptic neuron $\mathrm{j}$ is given by: 
\begin{equation} \label{eq8}
\mathrm{I_j} = \sum_{i=1}^n \mathrm{G_{i,j}} \mathrm{V_i} 
\end{equation}
where, $\mathrm{n}$ is the total number of rows, i.e. total number of pre-synaptic neurons. Note, as conductance values are always positive, in order to represent both negative and positive weights, each weight is actually mapped to a pair of PCM devices ($\mathrm{G_+}$,$\mathrm{G_-}$, shown in Figure \ref{crossbar} inset) where the positive (negative) conductance encodes the positive (negative) weight value while the other conductance is set to a high OFF resistive state. For sake of simplicity, in our following discussions, we will consider the conductance to be given by $\mathrm{G}=\mathrm{G_+} - \mathrm{G_-}$. The intrinsic device physics of PCM devices can be exploited to implement synaptic learning rules like Spike Timing Dependent Plasticity (STDP) \cite{sebastian2018tutorial} while the array structure enables the area and energy-efficient implementation of synaptic dot-product computation required in artificial neural networks \cite{fong2017phase}. 

However, memristive devices suffer from several non-idealities. While previous works on astrocyte enabled self-repair in neural networks \cite{wade_self-repair_2012, liu2017spanner,liu2018exploring,rastogi_self-repair_2021} have considered stuck-at-faults, hardware realistic faults are much more complex. In this work, we consider two specific hardware non-idealities: stuck-at-fault and weight drift. The stuck-at-fault is a type of fault where a pre-determined percentage of synapses are faulty or disabled, i.e. PR $=0$, and no longer affected by any dynamics. The weight drift scenario occurs in the healthy synapses and is representative of PCM device technology. Therefore, along with accounting for a portion of the synapses to be disabled (representing the simulation of stuck-at-faults) in our modelling, the PRs of healthy synapses drift with a randomly generated drift ratio. The drift fault pattern is extracted from the temporal PCM device characteristics \cite{ambrogio_reducing_2019}. The temporal decay of the conductance of a PCM device can be described as:
\begin{equation} \label{eq9}
\mathrm{G} = \mathrm{G_0} t_{\mathrm{norm}}^{-v}
\end{equation}
where, $\mathrm{G_0}$ is the initial conductance, $t_\mathrm{norm}$ is the normalized time, which is greater than 1, and $v$ is the log scale decay slope. The value $r_{\mathrm{decay}} = \frac{\mathrm{G}}{\mathrm{G_0}} = t_{\mathrm{norm}}^{-v}$ is called the drift ratio. Also, instead of a constant parameter, the log scale decay slope $v$ is sampled from a normal distribution for each individual PCM device. i.e.
\begin{equation} \label{eq10}
v ~ \sim \mathcal{N} (\mu_v, \sigma_v)
\end{equation}
where, $\mu_v$ and $\sigma_v$ represent the mean and standard deviation of the distribution of $v$. In both scenarios mentioned above, the fault is instantly injected to neuron N2 at time $t_{\mathrm{fault}} = $ 200 s. Neuron N1 is not affected by any fault. For the stuck-at-fault scenario, the impacted PRs drop to 0 at $t_{\mathrm{fault}}$ and remain at that value. For modelling both the stuck-at-fault and drift process, at $t_{\mathrm{fault}}$, the PRs of disabled synapses are set to 0 in the same way as the former. At the same time, each healthy synapse PR is multiplied by its individual $r_{\mathrm{decay}}$. In the following sections, without explicit declaration, the PRs are the PR values of neuron N2, as N1 is not influenced by fault.

\subsubsection{Self-Repair Macro-model Formulation}

In this subsection, we develop a macro-model capturing the temporal dynamics of the astrocyte induced self-repair process by utilizing the detailed computational model discussed previously. For simplicity, the macro-model is developed based on the response of the computational model to stuck-at-faults while performance evaluation on machine learning tasks (discussed in next section) is performed for both stuck-at-faults and weight drift non-idealities. Based on Equation \ref{eq6}, it can be observed that $\mathrm{PR}(t)$ is a product of $\mathrm{PR}(0)$ and a global variable $1 + (\mathrm{DSE}(t) + \mathrm{eSP}(t))/100$. Due to the stochastic nature of PR initialization and fault injection, the amount of self-repair varies in different simulation runs. In general, the PRs of healthy synapses increase after the occurrence of faults and finally stabilize at a certain level which is regulated by the competition between direct and indirect signal feedback. 
\begin{figure}[t]
\centering
\includegraphics[width=0.85\columnwidth]{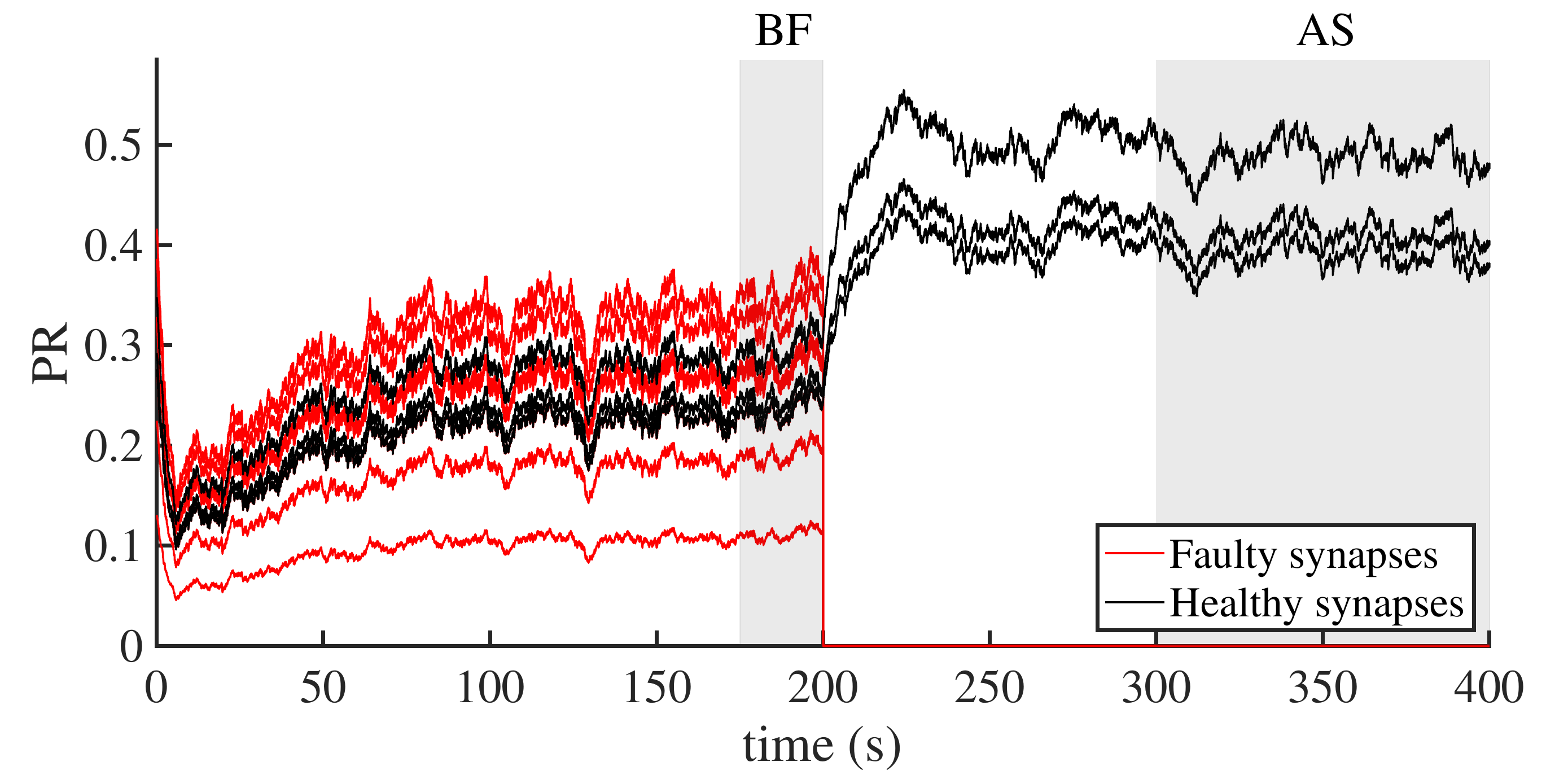}
\caption{An example of temporal evolution of PR value of neuron N2 under astrocytic influence, where the total number of synapses is 10.}
\label{self-repair-temporal}
\end{figure}
From Figure \ref{self-repair-temporal}, where a sample of the temporal evolution of PR values is plotted, it can be observed that the time cost for self-repair, which is defined as the time difference between $t_{\mathrm{fault}}$ and the time the PR takes to reach its repaired stable value, is similar for each synapse. In fact, we can prove that all the PRs keep their relative magnitude unchanged at any time during the simulation window unless there is any particular PR hitting the upper bound 1. At any given time $t_1$ and $t_2$, the ratio between two arbitrary PRs, say $\mathrm{PR}_i$ and $\mathrm{PR}_j$, is
\begin{equation} \label{eq11}
\begin{split}
    & \frac{\mathrm{PR}_i(t_1)}{\mathrm{PR}_j(t_1)} = \frac{\mathrm{PR}_i(0)(1 +  \frac{\mathrm{DSE}(t_1) + \mathrm{eSP}(t_1)}{100})}{\mathrm{PR}_j(0)(1 +  \frac{\mathrm{DSE}(t_1) + \mathrm{eSP}(t_1)}{100})}\\
    = & \frac{\mathrm{PR}_i(0)(1 +  \frac{\mathrm{DSE}(t_2) + \mathrm{eSP}(t_2)}{100})}{\mathrm{PR}_j(0)(1 +  \frac{\mathrm{DSE}(t_2) + \mathrm{eSP}(t_2)}{100})} = 
    \frac{\mathrm{PR}_i(t_2)}{\mathrm{PR}_j(t_2)}
\end{split}
\end{equation}
Therefore all the synapses finish their self-repair simultaneously. Also, from Equation \ref{eq11}, we can conclude that at any given time $t_1$ and $t_2$ and two arbitrary PRs, say $\mathrm{PR}_i$ and $\mathrm{PR}_j$, the following relation holds:
\begin{equation} \label{eq12}
\frac{\mathrm{PR}_i(t_1)}{\mathrm{PR}_i(t_2)} = \frac{\mathrm{PR}_j(t_1)}{\mathrm{PR}_j(t_2)}
\end{equation} 
In this work, the PR self-repair ratio $q$ of a synapse, say synapse $i$, is defined as the ratio of the PR value after self-repair and the stable PR value before fault injection:
\begin{equation} \label{eq13}
q_i = \frac{\overline{\mathrm{PR}_{i}(\mathrm{AS})}}{\overline{\mathrm{PR}_{i}(\mathrm{BF})}}
\end{equation}
where, $\overline{\mathrm{PR}_{i}(\mathrm{AS})}$ is the mean (averaged over multiple simulation runs) of stable PR value of synapse $i$ after self-repair and $\overline{\mathrm{PR}_{i}(\mathrm{BF})}$ stands for the mean of stable PR value of synapse $i$ just before fault injection. The intervals of AS and BF are shaded in Figure \ref{self-repair-temporal}. It is easy to infer from Equation \ref{eq12} that $q_1 = q_2 = .. = q_m$ where $m$ is the number of synapses of neuron N2. Therefore, we will denote $q$ without a subscript for neuron N2 for the remainder of the text.

\begin{figure}[t]
\centering
\includegraphics[width=0.85\columnwidth]{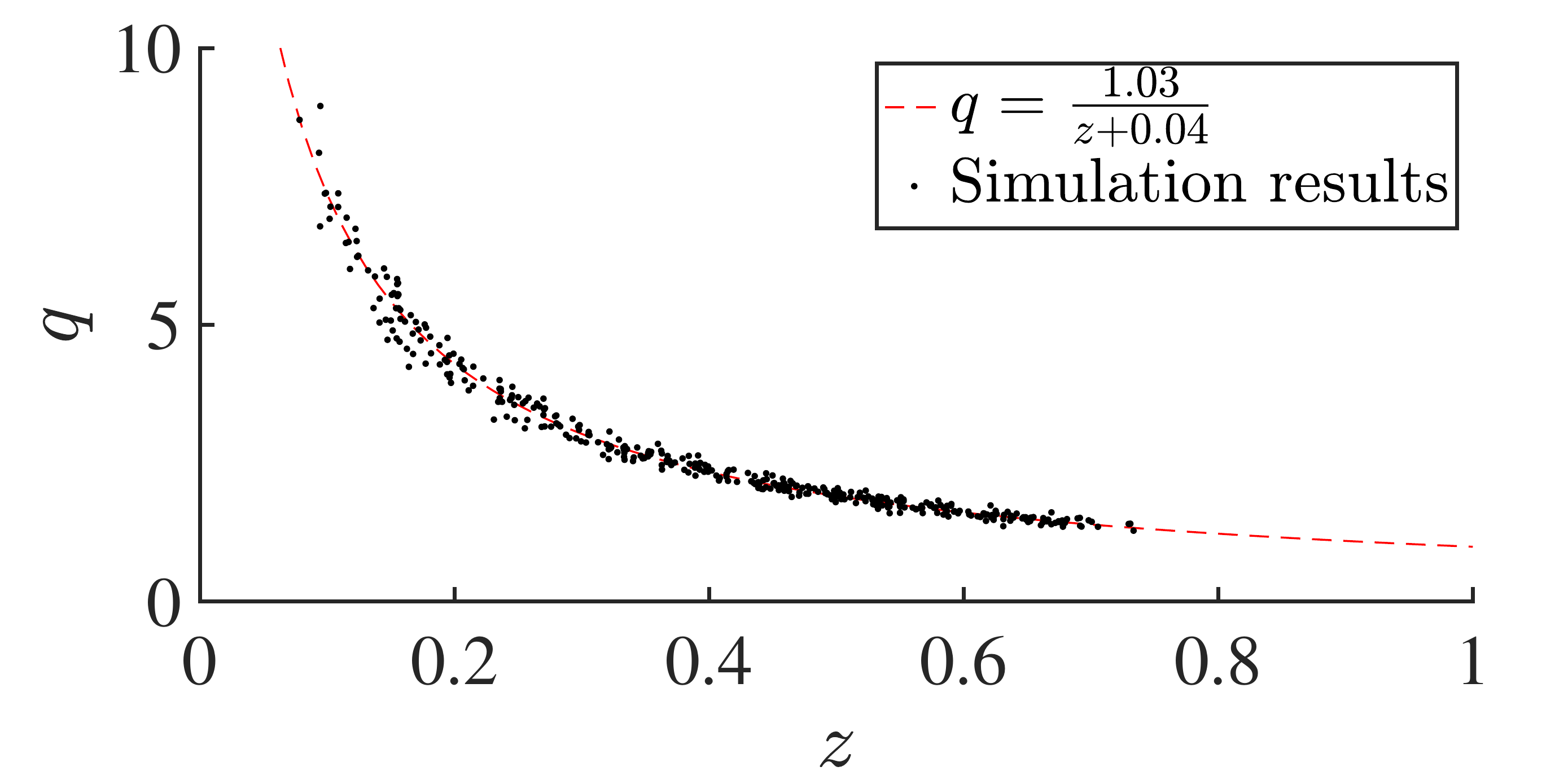}
\caption{Quantitative relationship between self-repair ratio, $q$, and fault severity, $z$.}
\label{q-z}
\end{figure}

Based on the computational model simulations, we inferred that the self-repair ratio $q$ is a function of the initial PRs and the fault magnitude. By statistics, the larger the percentage of affected synapses, larger will be the value of $q$. Based on data collected from 400 simulation runs with independent randomly sampled PR initialization and fault injection, it was confirmed that  $q$ is strongly correlated to the severity of the fault. Here, the severity of the fault $z$ is defined as:
\begin{equation} \label{eq14}
\begin{split}
    z & = \frac{\textrm{sum of initial PRs of non-disabled synapses}}{\textrm{sum of initial PRs of all the synapses}}\\
    & = \frac{\sum_{\textrm{non-disabled synapses}}\mathrm{PR}_i(0)}{\sum_{\textrm{all synapses}}\mathrm{PR}_i(0)}
\end{split}
\end{equation}
where, $\mathrm{PR}_i(0)$ is the initial PR of synapse $i$. In the context of terminology to be used in the next section for learning algorithm formulation in self-repair of neuromorphic hardware systems, a non-disabled synapse is a synapse whose PR is not stuck to zero and therefore includes synaptic weights with a drift factor. Quantitatively, the strong correlation found between $q$ and $z$ from the astrocyte computational model simulations is:
\begin{equation} \label{eq15}
q \approx \frac{1.03}{z + 0.04}
\end{equation}
which is also displayed in Figure \ref{q-z}.

For simplicity, an inverse proportion relation is considered in our algorithm design (to be discussed in next section), i.e.
\begin{equation} \label{eq16}
q \,\leftarrow\, \frac{1}{z}
\end{equation}
Such a strong correlation between $q$ and $z$ implies that the astrocyte self-repair mechanism will drive the PR of healthy synapses of a faulty neuron to increase until the sum of PR of the remaining healthy synapses approaches the original synaptic PR sum before faults. The temporal increment of PR of healthy synapses after fault injection follows an exponential-like trend since the self-repair becomes relatively slow as the PR of healthy synapses approach their target values. i.e.,
\begin{equation} \label{eq17}
\begin{split}
    \mathrm{PR}_i(t) & \approx  \overline{\mathrm{PR}_{i}(\mathrm{AS})}(1-e^{-\frac{(t-t_{\mathrm{fault}} + t_b)}{\tau}}) \\
    & = q \times \overline{\mathrm{PR}_{i}(\mathrm{BF})} (1-e^{-\frac{(t-t_{\mathrm{fault}} + t_b)}{\tau}}) \\[.4em]
    & \forall \,\, t > t_{\mathrm{fault}}
\end{split}
\end{equation}
where
\begin{equation} \label{eq18}
t_b = -\tau \log(\frac{q-1}{q})
\end{equation}
is the temporal intercept and $\tau$ is the self-repair time constant. The value of the temporal intercept can be determined from the initial condition that at $t = t_{\mathrm{fault}}$, the transmission probability starts to increase from an initial value $\overline{\mathrm{PR}_{i}(\mathrm{BF})}$.

The next section discusses astromorphic learning algorithm formulation for unsupervised SNNs inspired by the developed macro-model.

\subsection{Astromorphic Learning Algorithm Formulation}
\begin{figure}[h]
\centering
\includegraphics[width=0.85\columnwidth]{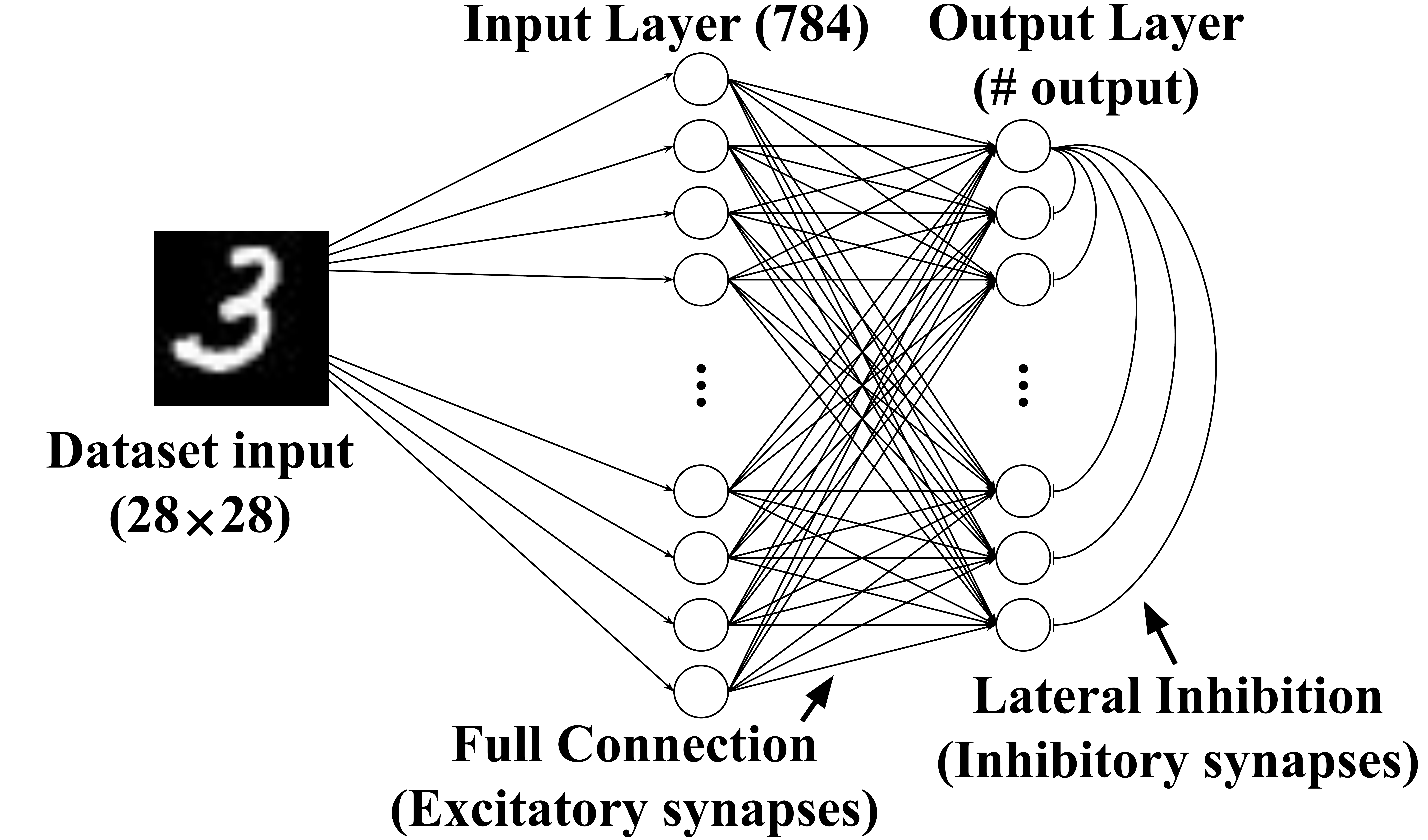}
\caption{SNN network architecture used for unsupervised learning. Lateral inhibitory connections are only shown for one neuron in the output layer.}
\label{stdp-net-arch}
\end{figure}
To demonstrate our proposal, we consider an SNN trained via the STDP learning algorithm where the synaptic weights are updated according to the timing of spikes of pre-synaptic and post-synaptic neurons. For detailed information about the learning algorithm and training process, interested readers are directed to \cite{diehl_unsupervised_2015}. An STDP network for image recognition has a structure as elaborated in Figure \ref{stdp-net-arch}. 
The dimension of the input layer, $n_{\mathrm{input}}$, is dictated by the dimension of the input images for recognition while the main network consists of $n_{\mathrm{neuron}}$ neurons.
To prevent single neurons from dominating the firing pattern during training, homeostasis is included in the neuron model: a neuron spike causes a constant threshold increment ($\theta^+$) over the baseline threshold. The threshold increment decays exponentially, according to decay time constant ($\tau_\theta$), with time when the neuron does not fire. Such a mechanism prevents a neuron from firing continuously, which enables the entire network to learn uniformly. The homeostasis effect is also balanced by a lateral inhibitory effect where any neuron spike in the output layer triggers negative spikes to all the other neurons (where the weights for all inhibitory connections is a global constant, $w_{\mathrm{inh}}$) and prevents them from firing, thereby promoting competitive learning. The representative counterpart of synaptic PR from the computational model is the set of synaptic weights in the SNN connecting the input and output neuron layers. There will be a positive weight update whenever the post-synaptic neuron fires after the pre-synaptic neuron and vice versa. The trace of a pre/post-synaptic neuron is set to 1 at the instant it generates a spike, and the trace decays exponentially afterward. The trace of a neuron measures the temporal closeness of the current time and its immediate previous spike. The STDP  learning rule based on post/pre-synaptic neuron spike traces can be formulated as:
\begin{equation} \label{eq19}
\Delta w(t) = \begin{cases}
    \eta_{post} \times x_{pre}(t) & \quad \Delta t > 0\\
    - \eta_{pre} \times x_{post}(t) & \quad \Delta t < 0\\
\end{cases}
\end{equation}
where, $\eta_{post}/\eta_{pre}$ are the post/pre-synaptic neuron learning rates and $x_{pre}/x_{post}$ are the traces of the pre/post-synaptic neurons. $\Delta t$ is the time difference between post-synaptic and pre-synaptic spikes, i.e. $\Delta t = t_{post} - t_{pre}$. Practically, the $\Delta t > 0$ case is executed when the post-synaptic neuron fires and the $\Delta t < 0$ case is executed when the pre-synaptic neuron fires.

Astrocyte feedback causes the modulation of synaptic plasticity in the presence of faults, in turn, modulating the STDP learning rule. Inspired by the inverse proportional relationship between the self-repair ratio and the fault ratio, the self-repair learning rule is expected to recover the original sum of synaptic PRs of a faulty neuron. The target repaired value of healthy synaptic weights of a faulty neuron is therefore determined by $q$. 
We note that an exponential increasing temporal pattern (exhibited by the dynamic repair process in astrocyte computational model in Equation (17)) is commonly represented by the following equation:
$\tau_x \frac{dx(t)}{dt} = x_{\mathrm{target}}  - x(t)$
, where $x_{\mathrm{target}}$ is the value that the objective $x$ converges to and $\tau_x$ is the convergence time constant. The slope of increment of the objective at any given time $t$ is proportional to the difference between the target value and the current value.
Therefore, the astrocyte modulated STDP learning rule can be described by the following equation where the rate of weight change is directly proportional to the difference between the target synaptic PR and the current weight: 
\begin{equation} \label{eq21}
\Delta w(t) = \begin{cases}
    \eta_{post} \times x_{pre}(t) \times \frac{q w_0 - w(t)}{\tau} & \quad \Delta t > 0\\
    - \eta_{pre} \times x_{post}(t) & \quad \Delta t < 0
\end{cases}
\end{equation}

where, $w_0$ is the weight of the synapse before fault injection, $w(t)$ is the current weight, $\tau$ is the self-repair time constant (which can be combined with the learning rate as a single hyperparameter), $q$ is the self-repair ratio mentioned in the previous section. 

Here, the self-repair ratio $q$ is approximated by $\frac{1}{z}$, where $z$ is a function of the sum of the synaptic weights of the corresponding neuron. It is therefore worth pointing out here that the resultant computational model inspired learning rule is \textit{local} where the synaptic weight updates are dependent on the current weight (\textit{local to synapse}) and sum of weights of the corresponding neuron (\textit{local to neuron}). This enables our algorithm to be memristive hardware compatible (for instance, sum of weights of a particular neuron can be evaluated by applying all-one voltages along the rows of the crossbar array in Figure \ref{crossbar}) unlike prior proposals \cite{rastogi_self-repair_2021} where the temporal dynamics of the weight updates during self-repair are governed by \textit{global} parameters like percentile of the weight distribution of the entire network and are also lacking in bio-fidelity from modelling perspective. 

The term $q w_0$ in Equation \ref{eq21} is the self-repair target value of the corresponding weight. Note that $q_{\mathrm{eff}}$ is not constant with time, therefore the target value of each synaptic PR/weight can change during the re-training process. 

The next section evaluates the performance of our proposed astromorphic learning rule against prior proposals in unsupervised SNNs for hardware-realistic faults. For clarity, we will refer to the basic spike-timing dependent synaptic plasticity as STDP, astrocyte mediated plasticity in prior proposals \cite{rastogi_self-repair_2021} as A-STDP (global) and our proposal as A-STDP (local). 
\begin{table*}[t]
\small
\centering
\begin{tabular}{|p{7.8em}|c|c|c|c|c|c|c|c|}
    \hline
    \multirow{4}{5em}{Dataset} & \multirow{4}{2em}{$p_{\mathrm{fault}}$} & Acc. & Acc. & \# Steps  &
    Acc. & \# Steps & Acc. & \# Steps \\

    & & Norm. & STDP & STDP & A-STDP & A-STDP & A-STDP & A-STDP\\
    
    & & & & & (global) & (global) & (local) & (local)\\
    
    & & (\%) & (\%) & ($10^4$ steps) & (\%) & ($10^4$ steps) & (\%) & ($10^4$ steps)\\
    
    \hline
    \multirow{5}{7.8em}{MNIST\\(Baseline: 91.43\%)} & 0.5 & 27.61 (0.96) & 83.22 (0.39) & 8.8 (2.5) & 76.26 (0.61) & 7.7 (1.8) & 87.31 (0.34) & 7.9 (1.6)\\ 
    & 0.6 & 27.18 (0.16) & 79.92 (0.62) & 9.0 (1.9) & 73.62 (0.78) & 5.8 (1.2) & 84.89 (0.32) & 6.2 (2.1)\\ 
    & 0.7 & 27.02 (0.76) & 70.11 (0.90) & 7.3 (1.0) & 70.83 (0.50) & 6.9 (1.8) & 82.45 (0.51) & 7.4 (2.5)\\ 
    & 0.8 & 26.33 (1.16) & 51.19 (0.86) & 5.0 (1.2) & 66.93 (0.31) & 8.0 (1.7) & 77.68 (0.85) & 6.3 (1.3)\\ 
    & 0.9 & 27.68 (0.25) & 39.41 (1.56) & 0.7 (0.6) & 65.42 (0.86) & 9.3 (1.4) & 68.58 (0.55) & 9.3 (1.2)\\

    \hline
    \multirow{5}{7.8em}{Fashion MNIST\\(Baseline: 77.60\%)} & 0.5 & 32.40 (1.42) & 75.75 (0.25) & 19.9 (4.9) & 67.30 (0.56) & 17.5 (3.4) & 76.14 (0.30) & 1.5 (0.9)\\ 
    & 0.6 & 32.18 (1.40) & 74.17 (0.47) & 21.4 (1.6) & 65.76 (0.12) & 15.1 (6.0) & 74.74 (0.23) & 2.5 (1.1)\\ 
    & 0.7 & 31.12 (0.38) & 72.63 (0.64) & 22.9 (0.8) & 64.49 (0.40) & 18.0 (4.8) & 72.67 (0.53) & 2.1 (1.1)\\ 
    & 0.8 & 30.30 (0.65) & 67.38 (0.58) & 19.5 (2.7) & 61.79 (1.01) & 19.6 (5.0) & 69.76 (0.43) & 2.2 (1.6)\\ 
    & 0.9 & 29.90 (0.38) & 43.48 (1.65) & 0.7 (0.6) & 57.75 (0.53) & 23.4 (0.7) & 64.54 (0.69) & 4.3 (2.6)\\

    \hline
\end{tabular}
\caption{Self-repair accuracy and convergence speed comparison among STDP, A-STDP (global) and our proposed A-STDP (local) learning rules for 400 neuron network trained on MNIST and Fashion-MNIST datasets. Statistics is averaged over 5 runs.}
\label{table1}
\end{table*}
\begin{table*}[t]
\small
\centering

\begin{tabular}{|p{7.5cm} p{1.1cm} | p{6.5cm} p{1cm}|}
\hline
Parameters & Values & Parameters & Values\\
\hline
Simulation time duration per image, $time$ & 100 ms & No. of output neurons, $n_{\mathrm{neuron}}$ & 400 \\

Simulation time-step size, $\Delta t$ & 1 ms & No. of input neurons, $n_{\mathrm{input}}$ & 784 \\

Membrane potential decay time const., $\tau_v$ & 100 ms & Batch size & 16 \\

Refractory period, $\delta_{\mathrm{ref}}$ & 5 ms & Spike trace decay time const.  & 20 ms \\

Resting membrane potential, $v_{\mathrm{res}}$ & -65 mV & STDP weight normalization factor & 78.4 \\

Reset membrane potential, $v_{\mathrm{reset}}$ & -60 mV & Maximum input spike rate (MNIST) & 128 /s \\

Threshold membrane potential, $v_{\mathrm{th}}$ & -52 mV & Maximum input spike rate (Fashion MNIST) & 45 /s \\ 

Adaptive threshold increment, $\theta_{+}$ & 0.05 mV & Inhibitory synaptic weight (MNIST), $w_{\mathrm{inh}}$ & -120 \\ 

Adaptive threshold decay time const., $\tau_{\theta}$ & 10$^7$ ms & Inhibitory synaptic weight (Fashion MNIST), $w_{\mathrm{inh}}$ & -250 \\ 

A-STDP (global) weight-percentile, $\alpha$ \cite{rastogi_self-repair_2021} & 98 & Post-synaptic learning rate (MNIST), $\eta_{post}$ & 10$^{-2}$ \\

A-STDP (global) non-linearity, $\sigma$ \cite{rastogi_self-repair_2021} & 2 & Post-synaptic learning rate (Fashion MNIST), $\eta_{post}$ & 4$\times$10$^{-3}$ \\

A-STDP (local) self-repair time const. (MNIST), $\tau$ & 10$^{-2}$ & Pre-synaptic learning rate (MNIST), $\eta_{pre}$  & 10$^{-4}$ \\

A-STDP (local) self-repair time const. (Fashion MNIST), $\tau$ & 4$\times$10$^{-3}$ & Pre-synaptic learning rate (Fashion MNIST), $\eta_{pre}$ & 4$\times$10$^{-5}$ \\

A-STDP (local) normalization period & 1 batch & Weight drift normalized time, $t_{\mathrm{norm}}$ & 10$^4$ \\

Weight sum lower bound (MNIST), $\mathrm{LB}_\mathrm{WeightSUM}$ & 0.17 & Weight drift log slope mean, $\mu_{v}$ & $1$ \\

Weight sum lower bound (Fashion MNIST), $\mathrm{LB}_\mathrm{WeightSUM}$ & 0.22 & Weight drift log slope standard derivation, $\sigma_{v}$ & 0.2258 \\

\hline
\end{tabular}
\caption{Self-repair retraining hyperparameters.}
\label{table2}
\end{table*}
\section{Results}
\subsection{Experiment Setup}
The proposed A-STDP (local) learning rule is implemented using the BindsNET \cite{hazan_bindsnet_2018} framework - an open-source SNN simulation platform based on PyTorch (https://pytorch.org/). The algorithms were run on the hardware environment consisting of one Intel(R) Xeon(R) Silver 4210 CPU, one NVIDIA GeForce RTX 2080 Ti GPU with 11264 MBytes graphics memory, 187 GBytes RAM and CentOS Linux 7 operating system. The SNNs are trained using STDP on two recognition tasks, namely the MNIST \cite{lecun_mnist_2010} and Fashion MNIST \cite{xiao_fashion-mnist_2017} datasets. The intensity of the pixels of the input images are converted into Poisson spike trains (the firing rates for different datasets are noted in Table II). The weights between the input layer and the output layer are randomly initialized with a uniform distribution of 0-0.3 range. Two types of hardware realistic faults (stuck-at-zero and weight drift) mentioned previously are injected in the baseline trained networks. Performance assessment in terms of accuracy and repair speed is performed for three generations of self-repair learning rules - STDP, A-STDP (global) \cite{rastogi_self-repair_2021} and A-STDP (local).

First, the baseline networks (which do not possess any fault) are prepared by training a randomly initialized network with the STDP learning rule. The baseline networks, consisting of 400 neurons, are individually trained on MNIST/Fashion MNIST dataset with a final accuracy of 91.43\%/77.60\%. Next, stuck-at-faults are injected in the weights of the baseline network with a certain fault probability $p_{\mathrm{fault}}$. i.e. each weight in the baseline network is set to zero with probability $p_{\mathrm{fault}}$. Five values for $p_{\mathrm{fault}} =$ 0.5, 0.6, 0.7, 0.8 and 0.9 are tested to mimic varying fault severity. The weight drift involves three parameters: $t_{\mathrm{norm}}$, $\mu_v$ and $\sigma_v$ (mentioned in prior sections).  Subsequently, STDP, A-STDP (local) and A-STDP (global) self-repair re-training is performed for comparative performance assessment. The best accuracy and number of data samples consumed before reaching best accuracy are recorded and analyzed. Interested readers can find the network weight maps before and after the introduction of non-idealities (stuck-at-faults and weight drift), including the self-repaired weights, in the supplementary material \cite{Han_Z_astromorphic_2022}. The detailed information regarding all simulation hyperparameters is included in Table \ref{table2} and typical ablation studies for important hyperparameters of the astromorphic learning model are also included in the supplementary material \cite{Han_Z_astromorphic_2022}.

Like previous works \cite{rastogi_self-repair_2021}, the weights of all neurons are normalized during the training process:
\begin{equation} \label{eq23}
\mathrm{WeightSUM}_i \leftarrow \overline{\mathrm{WeightSUM}}
\end{equation}
where, $\mathrm{WeightSUM}_i$ represents the sum of weights of neuron $i$ and $\overline{\mathrm{WeightSUM}}$ is the network-wide average of the sum of weights for all the neurons.
The normalization operation is applied after each batch of STDP learning, which guarantees that all the neurons have the same chance to be trained. Since the STDP algorithm updates the weights only when there is any spike and highly severe faults can result in very low spiking activity in the network, a lower bound of $\overline{\mathrm{WeightSUM}}$, namely $\mathrm{LB}_\mathrm{WeightSUM}$, is configured as a network hyperparameter to improve the training efficiency. 

\subsection{Accuracy and Convergence Speed Analysis}
Table 1 reports the best accuracy recorded in the self-repair process for hardware-realistic faults in the network (stuck-at-faults and weight drift). The results have been averaged over 5 independent runs of the network. In Table 1, Acc. Norm represents the network accuracy after fault injection and weight normalization. Acc. STDP/Acc. A-STDP (global)/Acc. A-STDP (local) is the accuracy measured after re-training using STDP/A-STDP (global)/A-STDP (local) learning rules respectively. \# Steps describes the number of training samples consumed before reaching the maximum accuracy during the self-repair process. The standard deviation of each measurement is included in the parentheses. A-STDP (local) significantly outperforms A-STDP (global) in terms of self-repaired accuracy. Further, while the convergence speed of A-STDP (local) is comparable to A-STDP (global) in case of the MNIST dataset, it is significantly faster over the more complex Fashion-MNIST dataset. Training convergence graphs are also provided in the supplementary material \cite{Han_Z_astromorphic_2022}. 

\section{Discussion}
This work strengthens the foundation for enabling astromorphic self-repair of hardware realistic faults in a neuromorphic system by forging stronger correlations with astrocyte neuroscience models. The key distinguishing aspect of the work lies in the neuroscience-hardware-software co-design aspect that delves into learning algorithm formulation that is not only neuroscience inspired but also hardware compatible. Performance assessment is provided in unsupervised learning tasks for hardware realistic faults going beyond simple stuck-at-faults considered in prior works. 

Future work can focus on understanding other aspects of astrocyte functionality in the self-repair process. For instance, there is a pair of feedback pathways in the astrocyte computational model based on the firing rate of the corresponding neuron, whose details have not been fully explored in this work.  Furthermore, the modulation of inhibitory synapses through astrocytes \cite{kang_astrocyte-mediated_1998} for neuromorphic computing still need more research. From a machine learning perspective, unsupervised self-repair mechanisms need to be also scaled further to deeper networks and large-scale datasets.

\section{{Acknowledgments}}
The work was supported in part by the National Science Foundation grants BCS \#2031632, ECCS \#2028213 and CCF \#1955815 and by Oracle Cloud credits and related resources provided by the Oracle for Research program.

\newpage

\section{Supplementary Material}

\subsection{Self-repair Time Constant Hyperparameter Tuning}
\begin{figure}[h]
\centering
\includegraphics[width=\columnwidth]{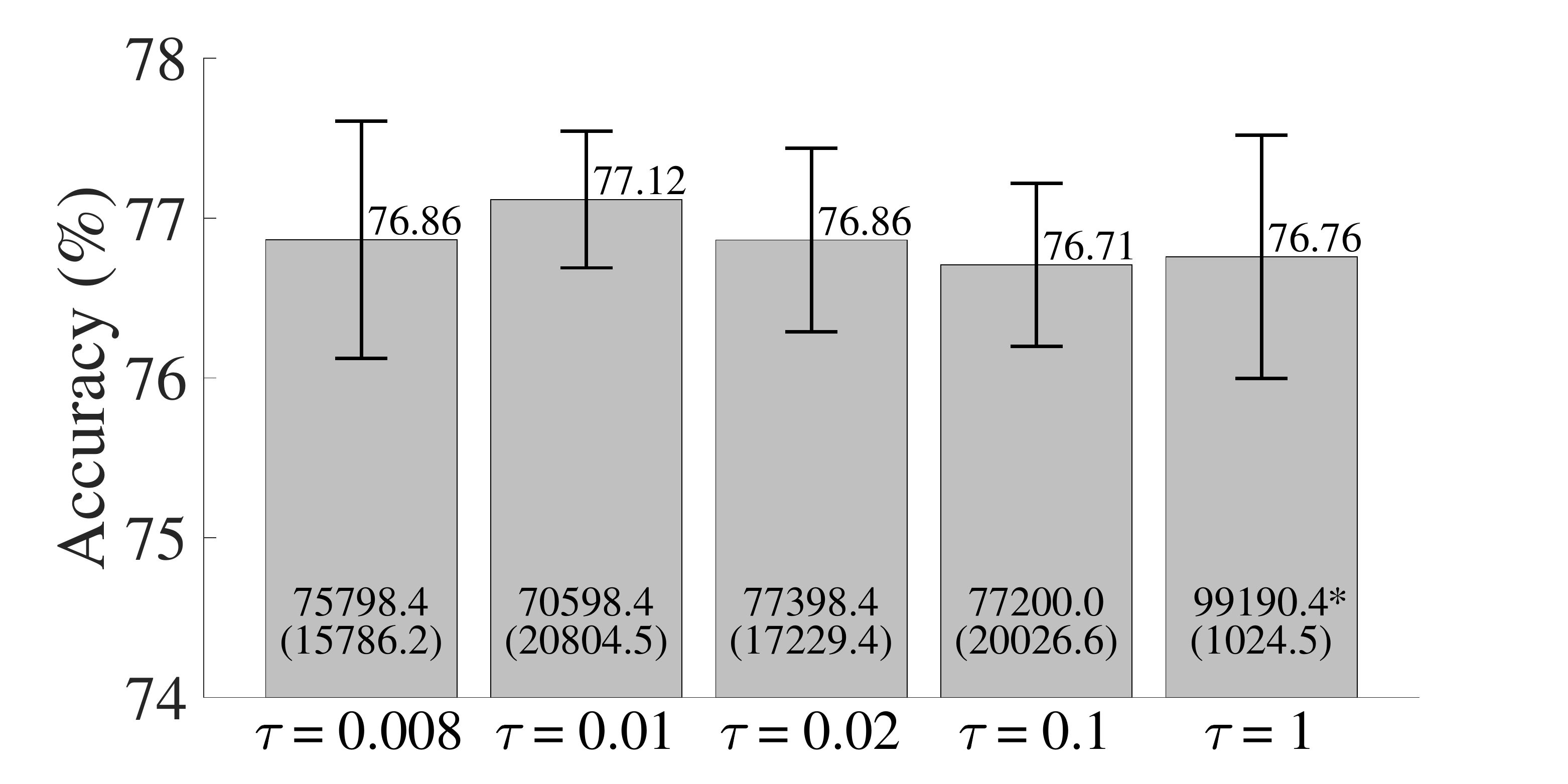}
\caption{Mean self-repaired accuracy with standard deviation (statistics from 10 runs) of A-STDP (local) algorithm retrained for 2 epochs on MNIST training dataset with different $\tau$ settings, where $p_{\mathrm{fault}}$ is 0.8 including weight drift. Mean accuracy value of each setting is shown on top of the bar. The lower portion of the bar shows the mean and standard deviation of the number of training samples required to reach optimal accuracy.}%\\$\ast$  At $\tau=1$, 2 out of 10 runs reach the optimal accuracy at the end of 2 epochs, implying insufficient convergence.}
\label{param-tune}
\end{figure}

While the majority of hyperparameter choices in the algorithm simulations were guided by prior literature, the self-repair time constant $\tau$ in the proposed A-STDP (local) learning rule had a significant effect on the convergence speed.
%First, the upper limit for weights should not be applied in self-repair retraining. As indicated in the results section, the sum of weights is an important factor determining the likelihood of firing of a neuron. A highly corrupted network needs a large self-repair ratio which will be prevented by a small upper limit of weights. In the experiment setting of this paper, 1000, which is reasonably large, is selected as an upper limit of all the weights in the network. In the experiments, there was no synapse weight \textcolor{red}{reaching} 1000 in any case, as the global weight sum for all the \textcolor{red}{neurons} is always close to 78.4, the uncorrupted baseline weight sum.
$\tau$ also controls the amount of weight update in a single repair step. %Additionally, $\tau$ is effective jointly with the post-synaptic neuron learning rate, as shown in Equation 20. 
The larger is the value of $\tau$, the slower the convergence speed will be. Ablation studies involving several values of $\tau$ were performed on the MNIST dataset where the self-repaired accuracy and mean and standard deviation of samples required to reach optimal accuracy are shown in Figure \ref{param-tune}. The MNIST training set (60000 samples in total) was randomly split into 50000 samples for training and 10000 samples for validation in order to perform the ablation studies. 10 individual runs were used to extract the statistics of Figure \ref{param-tune}.

%It should be noticed that $\tau=1$ is evaluated as a non-optimal setting because there is a probability of insufficient convergence, shown in Figure \ref{param-tune}. The other 4 settings of $\tau$s do not show a significant difference in terms of both accuracy and sample to optimal. Note that 0.01 is adopted as the learning rate for MNIST, which implies that in the condition of $\tau=0.01$, a healthy synapse could have its weight increased to the target value in a single retraining step before equalization.
Based on the above simulation studies, $\tau=0.01$ is selected as the optimal setting for the self-repair time constant hyperparameter.
%retraining of all fault severity, due to its relatively high accuracy and a small number of samples before reaching optimal. The learning rule declared in \cite{rastogi_self-repair_2021} diverges the weight where the originally larger weights will be even larger after A-STDP (global) self-repair, and vice versa. However, in drifted case, such a weight divergence strategy will fail because the relative magnitude among weights has been lost. When dealing with non-drift cases, weight divergence could be applied for further accuracy improvement.

\subsection{Weight Maps}
\begin{figure}
\centering
\begin{subfigure}{0.3\textwidth}
\includegraphics[width=\columnwidth]{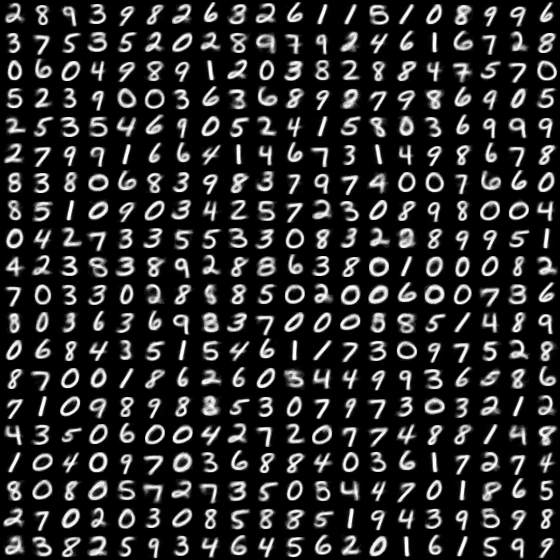}
\caption{Network with healthy weights}
\label{health}
\end{subfigure}
\hfill
\begin{subfigure}{0.3\textwidth}
\includegraphics[width=\columnwidth]{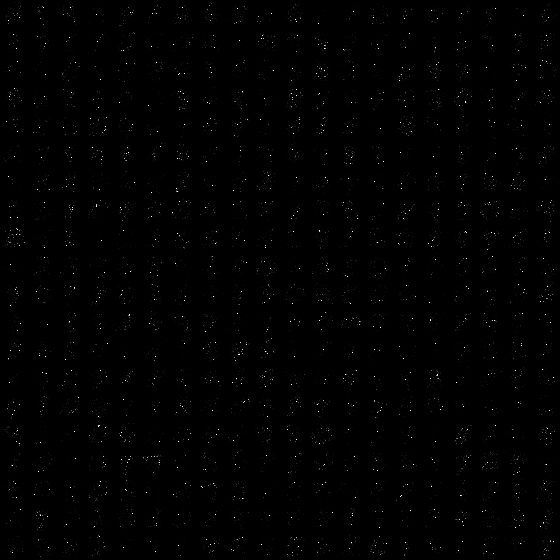}
\caption{Network after fault injection, weight drift, and normalization}
\label{after-norm}
\end{subfigure}
\hfill
\begin{subfigure}{0.3\textwidth}
\includegraphics[width=\columnwidth]{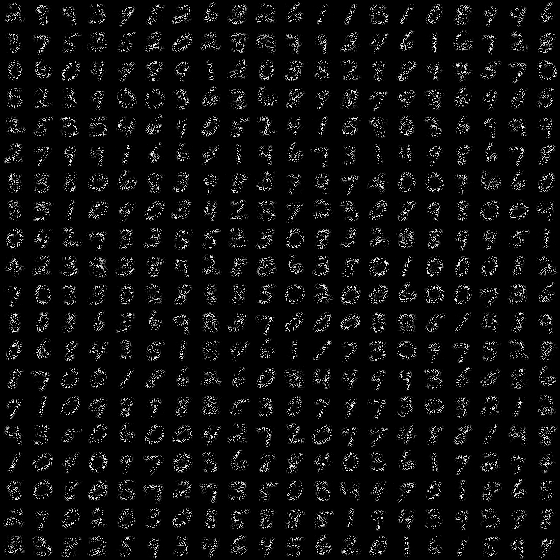}
\caption{Network after self-repair}
\label{after-repair}
\end{subfigure}

\caption{Weight maps of all neurons before and after faults/self-repair with $p_{\mathrm{fault}}=0.7$ including weight drift. For better illustration, weights are normalized to $[0,1]$ individually for each neuron.}
\label{weight-maps}
\end{figure}

Plotting the neuronal weights of the network as images intuitively illustrates the effect of faults and self-repair. Figure \ref{weight-maps} shows how the synapses' PRs are repaired after the injection of faults and weight drift, including the reference of healthy weights trained by STDP algorithm. 
From Figure \ref{after-norm}, it can be observed that the non-idealities degrade the weight maps for each neuron significantly. Figure \ref{after-repair} shows that A-STDP (local) retraining process is able to recover the weight maps to a reasonable degree.

\subsection{Convergence Plots}
\begin{figure}[h!]
\centering
\includegraphics[width=\columnwidth]{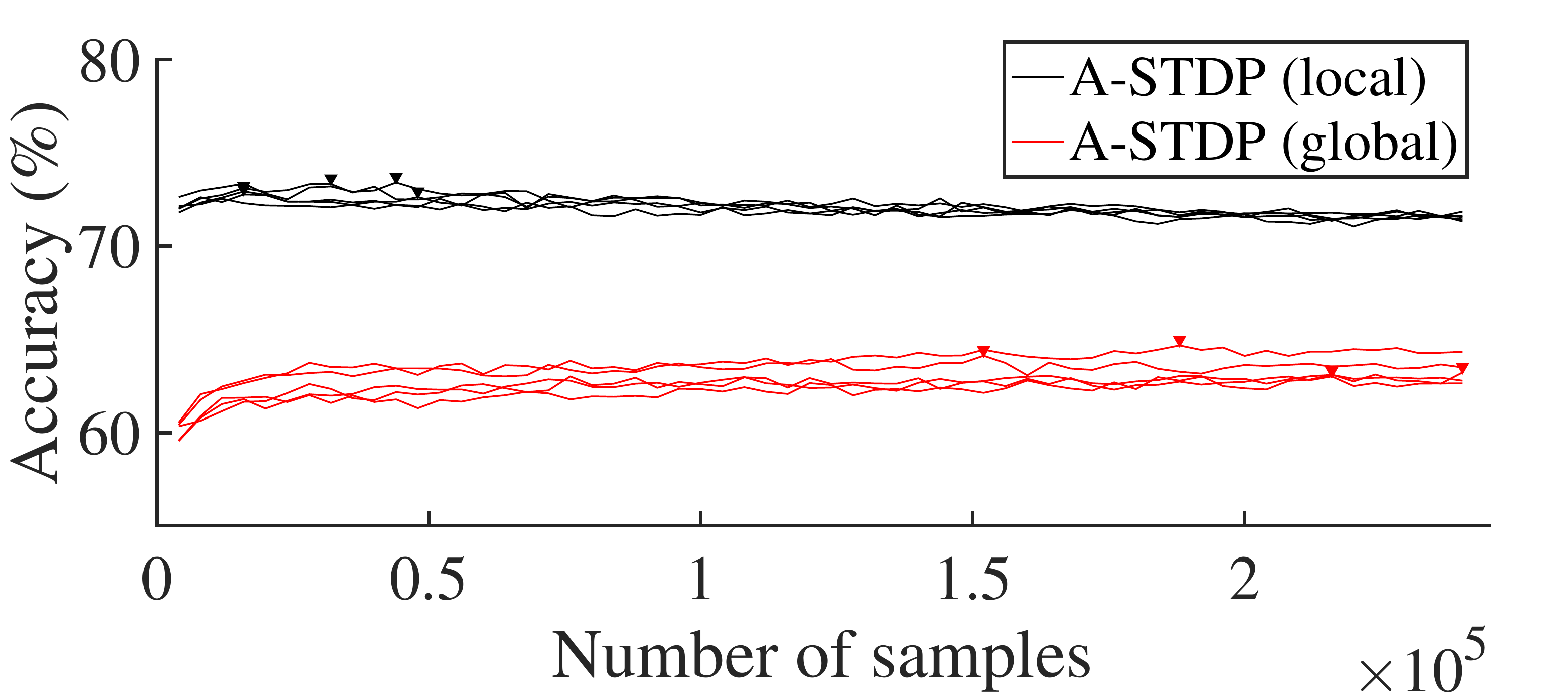}
\caption{Accuracy of 5 runs of A-STDP (global) and A-STDP (local) re-training processes for 4 epochs on Fashion MNIST dataset, where $p_{\mathrm{fault}}$ is 0.7 along with weight drift. The downward-pointing triangles mark the number of samples required for the algorithm to converge to the optimal accuracy for each run.}
\label{converge-plot}
\end{figure}

Convergence plots depicted in Figure \ref{converge-plot} show that the accuracy of A-STDP (local) reaches its maximum value during the re-training process within 50 thousand samples for Fashion MNIST dataset. However, A-STDP (global) converges much more slowly and the final accuracy is lower as well. 

\end{document}